\documentclass{article}

\usepackage{amsmath}
\usepackage[utf8]{inputenc}
\usepackage[T1]{fontenc}
\PassOptionsToPackage{numbers}{natbib}
\usepackage[final]{tackling_climate_workshop_style}
\usepackage{hyperref}  
\usepackage{multirow}
\usepackage{url}           
\usepackage{booktabs} 
\usepackage[normalem]{ulem} 
\usepackage{graphicx} 
\usepackage{float}
\usepackage{newunicodechar}
\DeclareUnicodeCharacter{2212}{\ensuremath{-}}
\usepackage{amsfonts}      
\usepackage{nicefrac}      
\usepackage{microtype}      
\usepackage{wrapfig}
\title{Mechanistic Interpretability of Atmospheric Rivers in GraphCast}

\author{
  Madelyn Mathai \\
 University of Virginia\\
  \And
    Timothy B. Higgins \\
    University of Virginia\\
  \And
   Kevin M. Grise\\
    University of Virginia\\
  \AND
  Chirag Agarwal\\
    University of Virginia\\
    \And
  Antonios Mamalakis\\
     University of Virginia\\
}

\begin{document}

\maketitle

\begin{abstract}
  While AI weather models now rival operational forecasts, how they represent the atmosphere internally remains an open question: feature attribution reveals which input patterns matter, not what the model computes or how it combines information internally. We train sparse autoencoders (SAEs) on GraphCast to uncover its learned concepts, using atmospheric rivers as our phenomenon of focus. Both standard and Matryoshka SAEs show GraphCast computes atmospheric river intensity, measured by integrated vapor transport (IVT), as a stable internal variable, despite IVT being neither an input nor a target. In contrast to the unstructured concept retrieval of the standard SAE, the Matryoshka SAE orders concepts by importance and exposes their relations. Atmospheric river concepts persist across depth and direct interventions confirm causality. This method offers a way to find internal variables and determine which of them the model actually relies on, which is a prerequisite for asking whether those variables remain meaningful as the phenomenon changes under a warming climate. Code available at \url{https://github.com/AikyamLab/climate_xai}.

\end{abstract}

\section{Introduction}
\looseness=-1 As deep neural networks (DNNs) are increasingly integrated into forecasting operations~\citep{benbouallegue2024rise, rasp2024weatherbench, lam2023graphcast}, it is crucial to understand how their predictions are made not only for assessing model accuracy, but also for enabling scientific discovery~\citep{mayer2024exploring}. While NWP systems are built on physical formulations and a shared theoretical foundation \citep{bauer2015quiet}, DNNs offer little to no understanding behind their decisions, hindering their adoption and scientific credibility~\citep{mcgovern2019making, ebert2020evaluation, doshi2017towards}. Prior work explains AI weather and climate models with post-hoc feature attribution, including relevance propagation, saliency, and perturbation methods \citep{bach2015pixel, toms2020physically, barnes2020indicator, sundararajan2017axiomatic, smilkov2017smoothgrad, ribeiro2016why, lundberg2017unified}, which reveal where a prediction is sensitive but not what the model has learned internally. Prior works that apply sparse autoencoders (SAEs) to climate science models only apply a standard SAE at a single depth~\citep{macmillan2025mechanistic}, serving as an implementation rather than a framework.

\looseness=-1 In this study, we focus on interpreting how GraphCast, an AI weather model, has learned to represent atmospheric rivers (ARs). We investigate how coherent internal representations, namely \emph{concepts}, can be identified in a network, how they are organized relative to each other, and how they can be traced across the network. To do this, we train two dictionary-learning architectures, Standard SAEs~\citep{gao2024scaling} and Matryoshka SAEs~\citep{bussmann2025matryoshka}, at layers 0, 8, and 15 within the model (Appendix~\ref{sec:saetrain}). 

\looseness=-1 We measure AR intensity with integrated vapor transport (IVT), a known function of GraphCast's humidity and wind inputs that lets us ground SAE latents in physical quantities. Our analysis shows that, although trained only to forecast and never on IVT, GraphCast computes AR intensity as a stable internal variable. The Matryoshka dictionary makes this concept not only detectable but also traceable in its relationship to other concepts. Direct intervention on the forward pass confirms the concept is causal, weakening real ARs across the globe when removed and intensifying their precipitation when written in. Because the approach extends to other phenomena and variables, this moves interpretability of weather models beyond input attribution toward a mechanistic account of what the model computes internally, how that computation is organized, and whether the forecast actually depends on it.

\section{Methods}

\begin{figure}
  \centering
  \includegraphics[width=\linewidth]{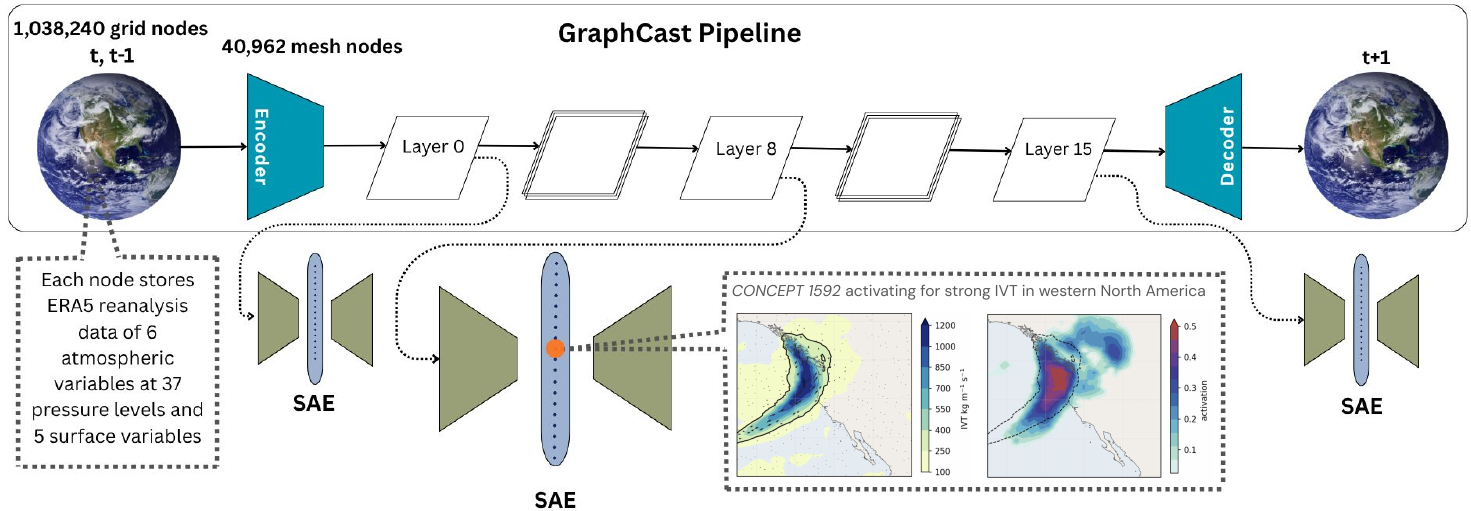}
  \vspace{-0.075in}
  \caption{\textbf{Top:} GraphCast maps ERA5 grid inputs at times $t-1$ and $t$ ($227$ variables per node) onto an icosahedral mesh, where a $16$-layer processor updates the $512$-dimensional mesh-node activations before the decoder returns the forecast at $t+1$. \textbf{Bottom:} We train SAEs post-hoc on the mesh-node activations at layers $0$, $8$, and $15$, each encoding the $512$-dimensional activation into a wider ($\mathbb{R}^{4096}$), sparsely active dictionary and reconstructing it. \textbf{Inset:} One of many AR concepts. The activation pattern of \textsc{Concept 1592} at layer $8$, which responds to strong IVT over western North America.}
  \label{fig:schematic}
\end{figure}
Fig. \ref{fig:schematic} shows GraphCast, a graph neural network (GNN) that autoregressively forecasts the global atmosphere at 0.25$^{\circ}$ resolution on a multiscale icosahedral mesh. GraphCast is a 16-layer GNN trained on ERA5 data from 1979 to 2017 at 6-hour time steps. The model predicts 227 fields spanning six atmospheric variables at 37 pressure levels and five surface variables, including 6-hourly accumulated precipitation. 

The Matryoshka SAE's ordinal dictionary allows us to explore how latents represent information relative to each other. The Matryoshka SAE prefix groups are $\text{G0}{=}[0,255]$, $\text{G1}{=}[256,511]$, $\text{G2}{=}[512,1023]$, $\text{G3}{=}[1024,2047]$, and $\text{G4}{=}[2048,4095]$, ordered from most to least important.  In the Matryoshka SAE, each prefix must reconstruct the input on its own, so the core fills with the most frequently active latents and the outer shells with progressively rarer, more specific ones. A broad latent cannot cede its coverage to a narrower one without hurting the earlier prefix's reconstruction \citep{bussmann2025matryoshka}. Matryoshka construction leaves behind the parent--child ordering itself \citep{kusupati2022matryoshka}: the latent from the earlier group can serve as the parent, and we can ask whether a later latent is a \emph{nested} child within it. Where language SAEs must infer relationships from activation co-occurrence alone \citep{bricken2023monosemanticity}, GraphCast supplies a handle that text does not: each latent activation has a spatial footprint on the mesh at each timestep, so containment becomes a geometric, physical question. We measure nesting as \emph{spatial containment},
$c = |\mathcal{F}^{\text{ch}} \cap \mathcal{F}^{\text{par}}| /
|\mathcal{F}^{\text{ch}}|$, the fraction of the child's (node, timestep) firings
at which the parent also fires, equivalently the confidence of the rule
``child $\Rightarrow$ parent'' \citep{agrawal1993mining}. \textbf{Our work is the first exploration of Matryoshka nesting structure in a GNN setting.} This allows us to ask questions like \emph{How do individual concepts work together to track atmospheric phenomena?} and \emph{Does a broad concept refine into narrower children by region or by intensity?}, questions that a flat, unordered dictionary cannot pose and that ground the relationships between latents in physical space rather than in co-occurrence alone. See Appendix~\ref{sec:structuralcontainment} for additional details on nesting.

\section{Results}
\subsection{Matryoshka SAE shows nesting structure for global ARs}

\begin{figure}
  \centering
  \includegraphics[width=\linewidth]{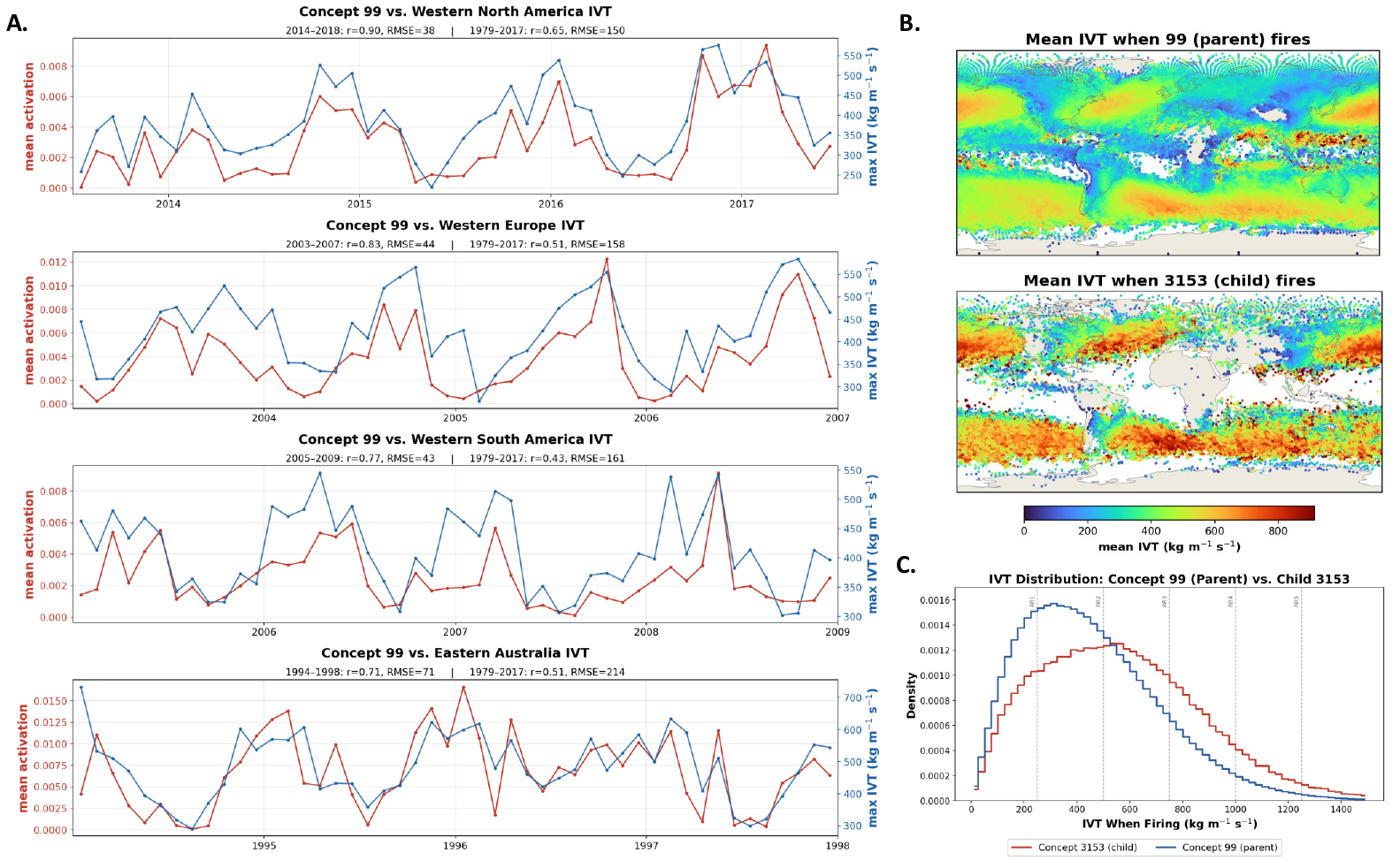}
  
  \caption{\looseness=-1\textbf{(A)} \textsc{Concept 99} tracks regional AR intensity in four regions among the most frequently affected by ARs
\citep{guan2015detection, waliser2017extreme, rutz2019artmip}: western North America ($30$--$50^\circ$N, $115$--$130^\circ$W), western Europe ($37$--$57^\circ$N, $8^\circ$W--$7^\circ$E), western South America ($30$--$50^\circ$S, $62$--$77^\circ$W), and eastern Australia ($22$--$42^\circ$S, $147$--$162^\circ$E). Displayed are region-mean activation of Matryoshka-L8 \textsc{Concept 99} (red, left axis) against region-maximum IVT (blue, right axis) over AR timesteps, averaged into monthly means. Each panel displays the four-year window with the strongest monthly correlation. Full training-period correlation is printed above each panel. \textbf{(B)} Every mesh node where \textsc{Concept 99} and child \textsc{Concept 3153} fire across 8{,}000 sampled timesteps, colored by the mean IVT at that node over the timesteps it fires;
blank regions are nodes where the concept never fires. \textbf{(C)} Distribution of node-level IVT over all sampled firing node--timesteps. Dashed lines indicate the AR category (Cat~1--5) thresholds based on magnitude alone; see \citet{ralph2019scale}.  } 
\label{fig:99fig}
\end{figure}

\textbf{Does a broad AR concept have more specific children?} \textsc{Concept 99} (core group G0) is the dictionary's strongest global AR concept, with a full-period mean correlation of $r=0.51$ between region-mean activation and region-max IVT across the four regions shown in Fig.~\ref{fig:99fig}A. Still, \textsc{Concept 99} fires diffusely, active at essentially every timestep and touching all but roughly $10\%$ of mesh nodes over the record (Fig.~\ref{fig:99fig}B). \textsc{Concept 3153} is contained in 99 at $\hat c = 0.953$ ($95\%$ CI $[0.949, 0.956]$, $n = 15{,}203$ firing events); 99's next-most-contained latent
reaches only $0.735$ ($[0.730, 0.739]$). No other latent in G0 or G1 contains
3153, so \textsc{Concept 99} is unambiguously its single parent.

\looseness=-1\textbf{What distinguishes the child's firings from the parent's?} Of 99's firing events, 3153 co-fires on the more intense ones. As shown in Fig.~\ref{fig:99fig}C, its firing sites carry higher IVT on average (mean $570$ versus $460~\mathrm{kg\,m^{-1}s^{-1}}$; $\Pr[\text{IVT}_{3153}>\text{IVT}_{99}]=0.61$), and the difference is sharpest in the extreme tail: 3153 fires on IVT above $750~\mathrm{kg\,m^{-1}s^{-1}}$ nearly twice as often as its parent ($26\%$ versus $14\%$). If the two distributions were completely non-overlapping, it would indicate \emph{feature absorption} \citep{chanin2024absorption}: \textsc{Concept 99} would have stopped firing on extreme ARs and ceded them to its child, creating a ``hole'' in the monosemantic concept it fires for (ARs). Instead, \textsc{Concept 3153} co-fires during the highest-IVT events but does not take over for \textsc{Concept 99}. The parent is present at $95\%$ of the child's firing sites, so 99 has not ceded these cases and is most active on them. It therefore nests by intensity: its child occupies the same geography but fires more on the strong end of the parent's condition.

\subsection{Steering shows causality of \textsc{Concept 99} }
\begin{wrapfigure}{r}{0.7\linewidth}
  \centering
  \includegraphics[width=\linewidth]{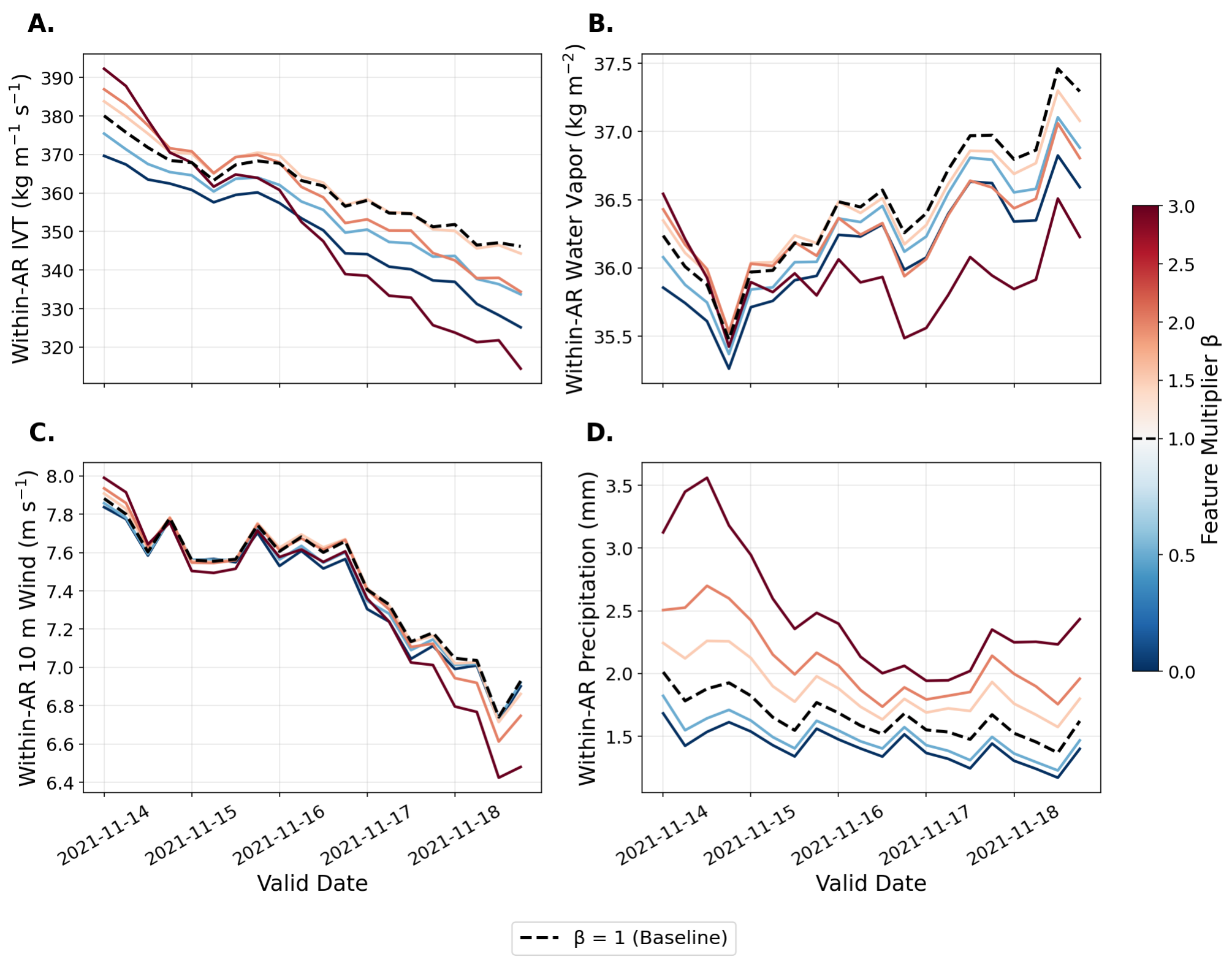}
  \vspace{-0.15in}\caption{\looseness=-1\textsc{Concept 99} dialed over a 5-day rollout of global ARs (initialized 13 Nov 2021, four years past the SAE training period). Panels show within-AR \textbf{(A)} IVT, \textbf{(B)} column water vapor, \textbf{(C)} $10\,\mathrm{m}$ wind, and \textbf{(D)} precipitation.}
  \label{fig:patching}
\end{wrapfigure}

\looseness=-1 To test causality, we scale a concept's SAE code by a factor $\beta$ at layer 8 and decode the change back into the mesh-node activations at every rollout step, so $\beta>1$ amplifies the concept, $\beta=1$ is the unedited baseline, and $\beta=0$ removes it (Appendix~\ref{sec:activation_steering}). As shown in Fig.~\ref{fig:patching}, dialing \textsc{Concept 99} moves within-AR precipitation strongly and monotonically while leaving wind nearly unchanged, so the concept acts on AR intensity through a moisture rather than a dynamical channel. 

\looseness=-1 IVT and column water vapor both increase with $\beta$ at initialization, but under strong amplification the enhanced precipitation depletes the column faster than transport resupplies it, and both fall below baseline within a day. Removing the concept instead weakens the AR uniformly, lowering IVT, vapor, and precipitation together. In contrast, applying the same causality test to \textsc{Concept 3153} produced no effect (not shown). GraphCast therefore holds concepts that only detect ARs, like \textsc{Concept 3153}, alongside those that drive them. Which of the many nested children are causal, and which merely detect, is a promising direction for future work.

\section{Conclusion and climate change impact}
\label{sec:conclusion}

Atmospheric rivers drive much of the world's extreme precipitation and flooding, and their intensity, the property warming is expected to change, is a quantity GraphCast was never given yet computes internally as a causal variable. 

\begin{wrapfigure}{r}{0.65\linewidth}
  \centering
  \includegraphics[width=\linewidth]{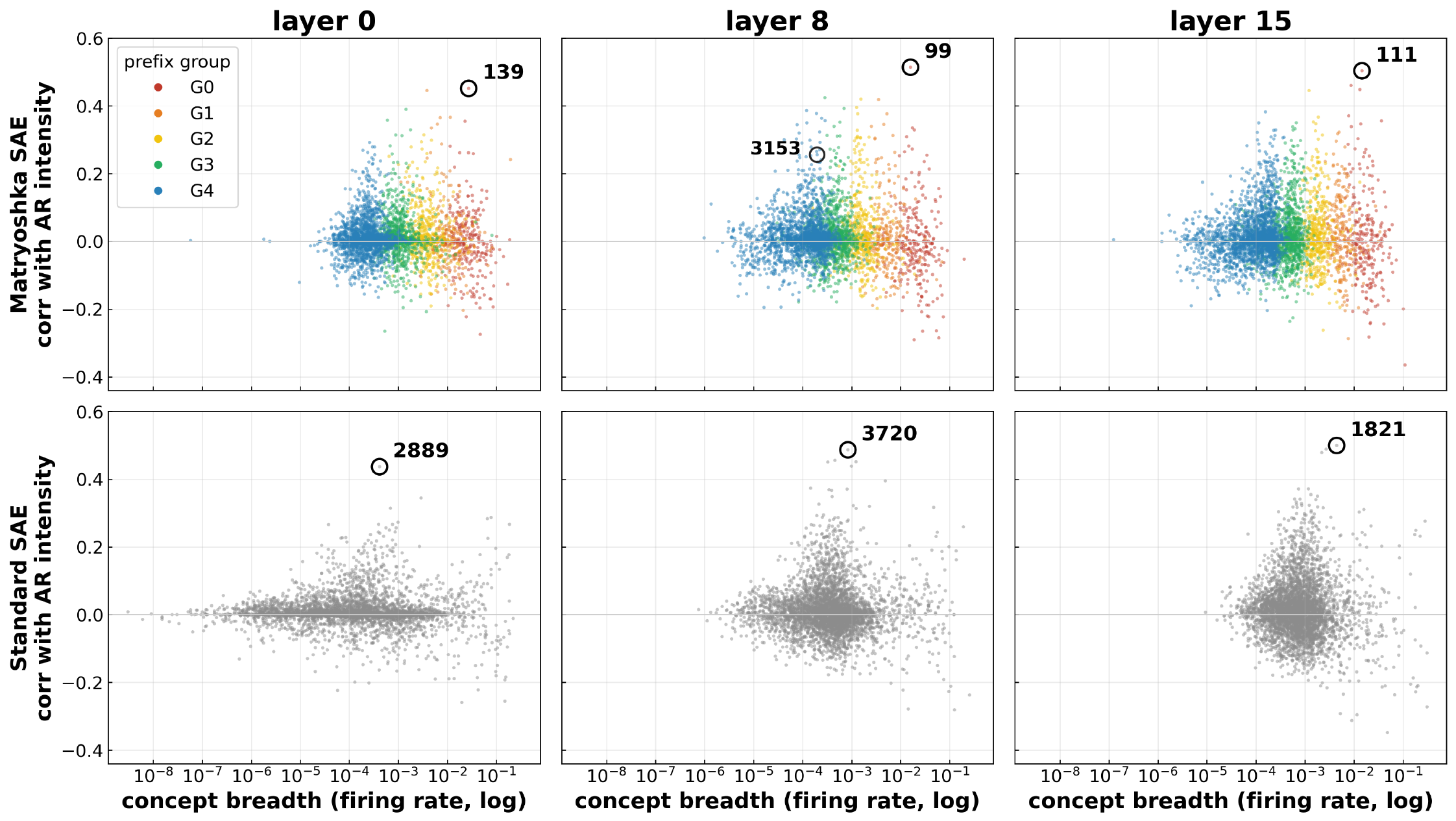}
  \caption{\looseness=-1 All $4{,}096$ concepts from each SAE plotted by breadth (firing rate) vs. correlation with AR intensity; Matryoshka SAE (top) and Standard SAE (bottom) at layers 0, 8, 15. Colors correspond to Matryoshka prefix groups. Each panel's top AR concept and child concept \textsc{Concept 3153} are labeled.}
  \label{fig:orgmap}
\end{wrapfigure}

Across depth the Matryoshka dictionary starts broad and refines, while the Standard SAE starts narrow and generalizes (Fig.~\ref{fig:orgmap}). The Matryoshka SAE gives that concept an address, holding it in the broad core at every depth while the standard SAE scatters the same function across arbitrary indices. This stable
addressing allows us trace a single concept across depth.

We characterized only one parent--child pair, though the sweep returned many more. Since its child \textsc{Concept 3153} detects the strongest ARs without
driving them, the full account of how GraphCast computes ARs lies in the circuit of related concepts rather than in any one of them. A model that computes a quantity it was never taught, is legible in a dictionary where scientists can locate it, and responds to intervention as the physics would suggest is no longer a black box but a system that can be audited as ARs intensify beyond its training record.
\clearpage
\bibliographystyle{unsrtnat}   
\bibliography{references}

\newpage
\appendix
\section{SAE Training}
\label{sec:saetrain}

\begin{table}
 \centering\small
 \setlength{\tabcolsep}{2pt}
 \renewcommand{\arraystretch}{1.25}
 \begin{tabular}{l c c}
 \hline
  Hyperparameter & Standard Top-$K$ & Matryoshka Top-$K$ \\
 \hline
   Input dimension $d$          & 512   & 512 \\
   Dictionary size $n$          & 4096  & 4096 \\
   Active latents $k$           & 32 (per sample) & 32 (per sample) \\
   Auxiliary latents (AuxK)     & 512   & 512 \\
   Input normalization          & mean-center, $\ell_2$ row & min--max $[-1,1]$, running-avg \\
   Batch size                   & 8192  & 4096 \\
   Training length              & 10 epochs & $3\times10^{5}$ steps \\
   Optimizer                    & Adam ($\beta{=}0.9, 0.999$; $\epsilon{=}6.25\times10^{-10}$)
                                & Adam ($\beta{=}0.5, 0.9375$) \\
   Learning rate                & $2\times10^{-4}$ & $2\times10^{-4}$ \\
 \hline
 \multicolumn{2}{l}{}
 \end{tabular}
\caption{\textbf{Training hyperparameters.} All three depths of a given architecture use identical settings; only the input activation record differs. Both architectures use per-sample Top-$K$ selection and an AuxK dead-latent revival term.}
 \label{tab:trainhp}
 \end{table}
\looseness=-1 All six SAEs share the same width and sparsity budget and differ only in input normalization, reconstruction objective, and optimization schedule (Table~\ref{tab:trainhp}). Each maps a $512$-dimensional GraphCast mesh activation to a dictionary of $4{,}096$ latents and keeps the $k=32$ largest per-sample pre-activations, so sparsity is fixed by construction and no $\ell_1$ penalty is used. Both architectures add an auxiliary Top-$K$ term (AuxK) that routes reconstruction error through the $512$ least-recently-active latents to revive dead latents. Optimization uses Adam at a learning rate of $2\times10^{-4}$ in every run.

\begin{wrapfigure}{r}{0.6\linewidth}
\centering
\includegraphics[width=\linewidth]{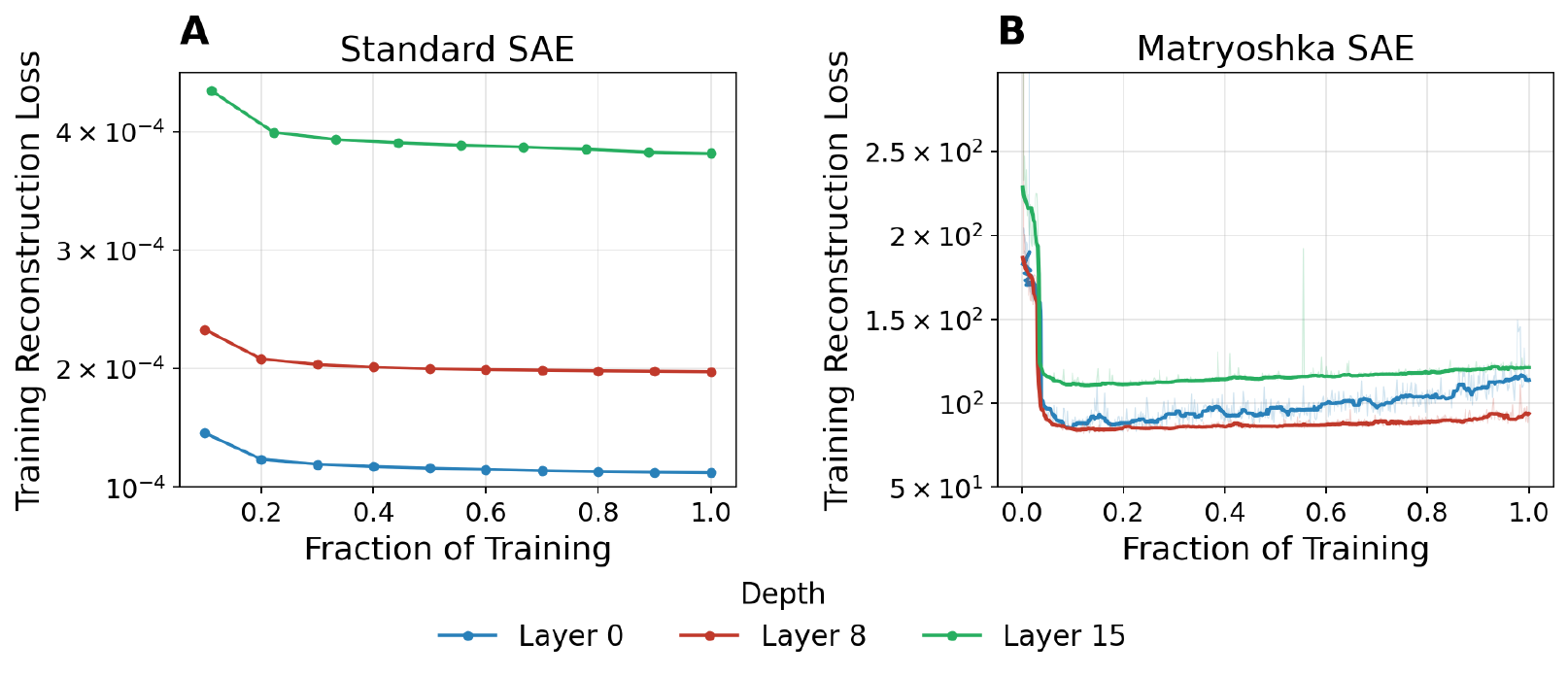}
\caption{Training reconstruction loss for the six SAEs against fraction of training, one panel per architecture, colored by depth. Each loss is measured in that SAE's own operating space, so the panels are not comparable in absolute value. The Matryoshka SAE logs every 500 steps, so its curve is drawn smoothed (bold) over the raw per-step values (faint) to show the trend.}
\label{fig:traincurves}
\end{wrapfigure}

The two architectures are trained on different schedules because their objectives and input normalizations differ. The Matryoshka loss averages reconstruction error over five nested prefixes, so each update carries gradient signal from five reconstructions rather than one, and its min--max input scaling and running-norm rescaling place activations in a different range from the Standard SAE's unit-normalized rows. Batch size, optimizer momentum, and schedule length were set separately for each so that each objective trains stably. As a result, the Standard SAE takes roughly $2.85\times10^{6}$ updates over $10$ epochs, while the Matryoshka SAE takes $3\times10^{5}$ updates covering about half an epoch. We treat each schedule as sufficient rather than matched: in both architectures the training loss plateaus well before the schedule ends, so additional updates would not materially change either dictionary.  

Figure~\ref{fig:traincurves} shows the reconstruction loss over training for all six SAEs, each in its own operating space. Each panel reports its architecture's native training loss. The Standard SAE is trained on unit-normalized activations and the Matryoshka SAE on min-max-scaled activations, so absolute loss values are not comparable across panels; only the trend within a panel is meaningful. The active-latent count remains fixed at k=32 throughout Matryoshka training, and the slight late rise in its loss reflects optimization under a constant learning rate against a running input normalizer rather than any loss of reconstruction quality.

\section{Matryoshka Nesting}
\label{sec:structuralcontainment}
As this is the first work to explore Matryoshka nesting in a GNN setting, we provide definitions that specify spatiotemporal relationships between concepts. Any two latents active somewhere on nearly every timestep will trivially co-fire at the timestep level, so timestep co-occurrence cannot distinguish nesting. Such latents are nonetheless active at only a small fraction of the nodes at any given moment. We therefore ask not whether parent and child fire on the same input, but whether they fire at the same nodes on that input. We take every (node, timestep) at which the child fires and ask how often the parent is firing there too. We call this fraction $c$ the child's \emph{spatial containment} in the parent. A latent is counted as firing at a node when its activation exceeds zero, \textit{i.e.}, it is one of the top-$k$ active latents at that node. Spatial containment is the \emph{confidence} of the association rule ``child $\Rightarrow$ parent'' \citep{agrawal1993mining}, the set-inclusion coefficient
\[
c \;=\; \frac{|\mathcal{F}^{\text{ch}}\cap\mathcal{F}^{\text{par}}|}{|\mathcal{F}^{\text{ch}}|},
\]
where $\mathcal{F}^{\text{ch}}$ and $\mathcal{F}^{\text{par}}$ are the sets of (node, timestep) cells at which the child and parent fire. Dividing by the child's own firing count is what makes the measure one-directional: the parent may fire at many nodes the child never touches without lowering $c$. Treating each of the child's firing events as a Bernoulli trial, we report $c$ as a proportion with a $95\%$ Wilson score interval. This relation is one we measure, not one the architecture imposes: the Matryoshka objective orders latents into nested groups but never binds a later latent to a unique earlier one, so containment induces a partial order rather than a hierarchy. A latent may nest under zero, one, or several parents.

We form every candidate pair by taking each latent in the two innermost groups
(G0 and G1, latents $0$--$511$) with every later-group latent ($1{,}835{,}008$
pairs) and compute $c$. For the overwhelming majority containment is essentially
zero (median $0.003$, with $90\%$ of pairs below $0.083$). Every one of these
$1.8$ million unrelated pairs is in effect a random pairing, so the sweep is its
own negative control. Genuine nesting is rare: just $7{,}995$ pairs ($0.44\%$)
exceed $c=0.5$ and only $1{,}436$ ($0.078\%$) reach $c=0.9$. These pairs are
asymmetric in the way nesting requires: the parent is nearly always present where
the child fires (high $c$), yet the child covers only a few percent of the
parent's own footprint. The two form a part and a whole rather than feature splitting, which would be two copies
of the same concept \citep{chanin2024absorption}. 

We call a later-group latent \emph{nested} in an earlier-group parent when the lower end of its $95\%$ Wilson interval exceeds $0.5$, so the child fires inside the parent more often than not; \emph{strongly} nested when the lower end of its $95\%$ Wilson interval exceeds $0.9$; and \emph{exclusively} nested when it is strongly nested in that parent \emph{and in no other latent in G0 or G1}. As containment is only a partial order, strong nesting need not be exclusive: a narrow latent lying inside the overlap of two broad concepts is strongly contained in both. The pair examined in Section 3.1 is both strong and exclusive.

\section{Activation Steering}
\label{sec:activation_steering}
Both the Standard SAE and Matryoshka SAE normalize model activations prior to SAE encoding. Each architecture does this slightly differently. The Standard SAE normalizes each activation vector separately. It centers it at the mean and rescales it to unit length, so the normalization factor is that vector's own magnitude and is different for every activation. Because this factor is known only during the forecast, the edit is converted back to the activation's original units, by multiplying by that same magnitude, inside the forward pass so the model can continue. The Matryoshka SAE instead normalizes each of the 512 activation dimensions by its fixed range over the training period, mapping each dimension's minimum to $−1$ and maximum to $+1$. Since those ranges are fixed constants, converting the edit back to original units uses a fixed factor, $m^{\max} - m^{\min}$, so the whole edit can be precomputed.

With the understanding of how each normalization works, we can write a concept edit. A concept $c$ contributes a fixed pattern to a node's activation vector $x$: its activation strength $z_c$, the code value the SAE assigns the concept, times its decoder direction $W_{\mathrm{dec}}[c]$, the pattern in activation space the concept stands for. To intervene, we change the concept's code by an amount $\delta z_c$ and add the decoded change $\delta z_c\,W_{\mathrm{dec}}[c]$ back into the activation, either scaling an already-active concept to $\beta$ times its value or writing a chosen level $b$ into a silent one. Because $W_{\mathrm{dec}}[c]$ is expressed in the SAE's normalized space, we convert the change to the activation's raw units with the same factor that undoes the normalization.

For the Standard SAE that factor is the node's own magnitude, the length
$\lVert x - \bar{x}\rVert$ of its mean-centered activation, where $\bar{x}$ is the mean
over the $512$ dimensions:
\begin{equation}
\Delta x_{sae} \;=\; \delta z_c \; W_{\mathrm{dec}}[c] \; \lVert x - \bar{x} \rVert,
\qquad
\delta z_c =
\begin{cases}
(\beta - 1)\, z_c & \text{scale an active concept by } \beta\\[4pt]
b & \text{write a value } b \text{ into a concept}
\end{cases}
\label{eq:cases}
\end{equation}
Since $\lVert x - \bar{x}\rVert$ is known only once the forecast produces the activation, this edit is applied inside the forward pass. For the Matryoshka SAE the factor is a fixed constant set by the per-dimension bounds $m^{\min}$ and $m^{\max}$, the smallest and largest value each dimension takes over training, and the scalar $s = \sqrt{d}/\bar{n}$, where $d = 512$ is the activation dimension and $\bar{n}$ the
running norm:
\begin{equation}
\Delta x_{matry} \;=\; \delta z_c \; W_{\mathrm{dec}}[c]\,
\frac{m^{\max} - m^{\min}}{2 s},
\qquad
\delta z_c =
\begin{cases}
(\beta - 1)\, z_c & \text{scale an active concept by } \beta\\[4pt]
b & \text{write a value } b \text{ into a concept}
\end{cases}
\label{eq:delta}
\end{equation}
so the whole edit is a fixed field, computed once over the training period and added to the activations during the forecast.

\end{document}